\documentclass[twoside,twocolumn]{article}

\usepackage{tabulary,graphicx,times,caption,fancyhdr,amsfonts,amssymb,amsbsy,latexsym,amsmath}
\usepackage{tabularx}
\usepackage{booktabs}
\usepackage{lscape}
\usepackage[utf8]{inputenc}
\usepackage{url, multirow,morefloats,floatflt,cancel,tfrupee,textcomp,colortbl,xcolor,pifont}
\usepackage[nointegrals]{wasysym}
\usepackage{rotating}
\usepackage{subfig}
\usepackage{svg}

\newcommand{\textBF}[1]{%
    \pdfliteral direct {2 Tr 0.3 w} 
     #1%
    \pdfliteral direct {0 Tr 0 w}%
}

\makeatletter

\usepackage{ifxetex}
\ifxetex\else
  \usepackage{dblfloatfix}
\fi

\@ifundefined{subparagraph}{
\def\subparagraph{\@startsection{paragraph}{5}{2\parindent}{0ex plus 0.1ex minus 0.1ex}%
{0ex}{\normalfont\small\itshape}}%
}{}

\def\URL#1#2{\@ifundefined{href}{#2}{\href{#1}{#2}}}

\def\UrlOrds{\do\*\do\-\do\~\do\'\do\"\do\-}%
\g@addto@macro{\UrlBreaks}{\UrlOrds}

\makeatother

\usepackage[paperheight=11in,paperwidth=8.3in,margin=2.5cm,headsep=.7cm,top=2.5cm]{geometry}
\usepackage[T1]{fontenc}

\renewenvironment{abstract}
	{\trivlist\item[]\leftskip0pt\par\vskip4pt\noindent
  	\textbf{\abstractname}\mbox{\null}\\}
	{\par\noindent\endtrivlist}

\def\keywords#1{\par\medskip\par\noindent\textbf{Keywords}: #1\par}

 \date{} \emergencystretch 8pt

\makeatletter
\def\author#1{\gdef\@author{\hskip-\tabcolsep%
	\parbox{\textwidth}{\raggedright\bfseries#1\\[1pc]}}}
\def\address[#1]#2{\g@addto@macro\@author{\\\hskip-\tabcolsep\parbox{\textwidth}{\raggedright%
	\normalsize\normalfont\textsuperscript{#1}#2}}}
\let\addresslink\textsuperscript
\def\correspondence#1{\g@addto@macro\@author{\\\hskip-\tabcolsep\parbox{\textwidth}{\raggedright%
	\vspace*{10pt}\normalsize\normalfont~\\#1~\\[12pt]}}}
\def\email#1{\g@addto@macro\@author{\\\hskip-\tabcolsep\parbox{\textwidth}{\raggedright%
	\normalsize\normalfont Emails: #1}}}

\def\title#1{\gdef\@title{\vspace*{-30pt}%
	\raggedright\textbf{\@journaltitle}~\\%
  \raggedright\bfseries\ifx\@articleType\@empty\vspace*{20pt}\else%
  \vspace*{20pt}\@articleType\vspace*{20pt}\\\fi#1}}
\let\@journaltitle\@empty \def\journaltitle#1{\gdef\@journaltitle{{\normalfont\itshape#1}}}
\let\@articleType\@empty \def\articletype#1{\gdef\@articleType{{\normalfont\itshape#1}}}

\let\@runningHead\@empty \def\RunningHead#1{\gdef\@runningHead{{\normalfont #1}}}

\usepackage{fancyhdr}
\fancypagestyle{headings}{\fancyhf{}
  \fancyhead[R]{\itshape\@runningHead}
  \fancyfoot[C]{\thepage}}
\fancypagestyle{plain}{%
	\fancyhf{}\fancyhead[R]{\LaTeX\ template}
  \fancyfoot[C]{\thepage}}
\makeatother

\usepackage[%
	numbers,sort&compress%
  ]{natbib}

\usepackage{float,xcolor}
\usepackage{amsmath}
\usepackage{graphics,graphicx}

\articletype{Research Article} 
\begin{document}

\title {Enhanced Multi-level Features for {Very High Resolution} Remote Sensing Scene Classification}

\author{%
		C Sitaula* \addresslink{},
            S KC \addresslink{} and
	      J Aryal+\addresslink{}
    }
		
\address[]{Earth Observation and AI Research Group, Department of Infrastructure Engineering, The University of Melbourne, Parkville, 3010, Victoria, Australia}

\correspondence{ Correspondence should be addressed to 
    	C Sitaula:  chiranjibi.sitaula@unimelb.edu.au}

\email{
chiranjibi.sitaula@unimelb.edu.au (C Sitaula * Corresponding author),
s.kc@unimelb.edu.au (S KC), 
jagannath.aryal@unimelb.edu.au (J Aryal+ Research Group Leader)
}%

\RunningHead{Running head}

\maketitle 

\begin{abstract}
Very high-resolution (VHR) remote sensing (RS) scene classification is a challenging task due to the higher inter-class similarity and intra-class variability problems. Recently, the existing deep learning (DL)-based methods have shown great promise in VHR RS scene classification. However, they still provide an unstable classification performance. To address such a problem, we herein propose a DL-based novel approach using an enhanced VHR attention module (EAM), which captures the richer salient multi-scale information for a more accurate representation of the VHR RS image during classification. Experimental results on two widely-used VHR RS datasets show that the proposed approach yields a competitive and stable/consistent classification performance with the least standard deviation of 0.001. Further, the highest overall accuracies on the AID and NWPU datasets are 95.39\% and 93.04\%, respectively. Such encouraging, consistent and improved results shown through detailed ablation and comparative study underscore the potential use of the proposed approach by the remote sensing community for the land use and land cover classification problems with more trust and confidence.
\end{abstract}

\section{Introduction}  

Remote sensing (RS) scene classification using very high-resolution (VHR) images, {which is a source of spatial information}, has attracted wider attention in robotics and earth observation. With the proliferation of such images, there has been rapid progress in VHR RS scene image interpretation tasks, such as scene classification \cite{dutta2023remote}, image retrieval, object detection, and semantic segmentation \cite{wang2023p}. Nevertheless, challenges such as complex spatial structures, diverse semantic categories, higher inter-class similarity, and intra-class variability have resulted in sub-optimal classification performance.

With the predominance of deep learning (DL)-based methods, particularly convolutional neural networks (CNNs) \cite{ouma2023flood,shen2023hla} with recent advances, the classification performance of VHR RS scene images has been improved these days as in indoor/outdoor scene image classification \cite{sitaula2021scene,sitaula2023recent}. The DL-based methods that leverage transfer learning in this domain have been focused on either an end-to-end approach \cite{wang2022} or a two-step approach \cite{Weng2017Land-UseFeatures}. The end-to-end approach exploits the fine-tuning strategy and employs either the uni-modal-based multi-level feature fusion \cite{wang2022} or the ensemble approach \cite{Scott2018EnhancedSets}.
Several works have been carried out in the literature using the end-to-end approach for the VHR RS scene classification.
For example, Wang et al. \cite{wang2018scene} employed the attention recurrent convolutional network (ARCNet) for the VHR RS scene classification that helped in selecting the salient spatial regions sequentially.
Furthermore, He et al. \cite{he2019skip} designed a skip-connected covariance (SCCov) network to classify the VHR RS scene images. 
The classification performance was improved based on the second-order information from the SCCov network.
Considering the object relationship importance, Li et al. \cite{9178726} established a DL model called object relationship reasoning CNN (ORRCNN), which adopts the scene and object-detection pipeline. Their work resulted in the overall improvement of the classification performance. 
Similarly, Wang et al. \cite{wang2022} developed a multi-level feature fusion (MLFF) module, which employs a novel adaptive channel dimension reduction approach and improved accuracy significantly.
Furthermore, Li et al. \cite{Li2022} devised a multi-scale residual network with higher-order feature information to improve the classification performance. Following the efficacy of multi-scale information in scene classification problems, Wang et al. \cite{9298485} recently devised the global-local two-stream architecture and significantly improved the VHR RS scene classification performance. 
Despite the significant improvement in the overall classification performance by this approach, it has less generalisability.

 \begin{figure*}
  \centering
  \includegraphics[height=85mm,width=0.75\textwidth, keepaspectratio]{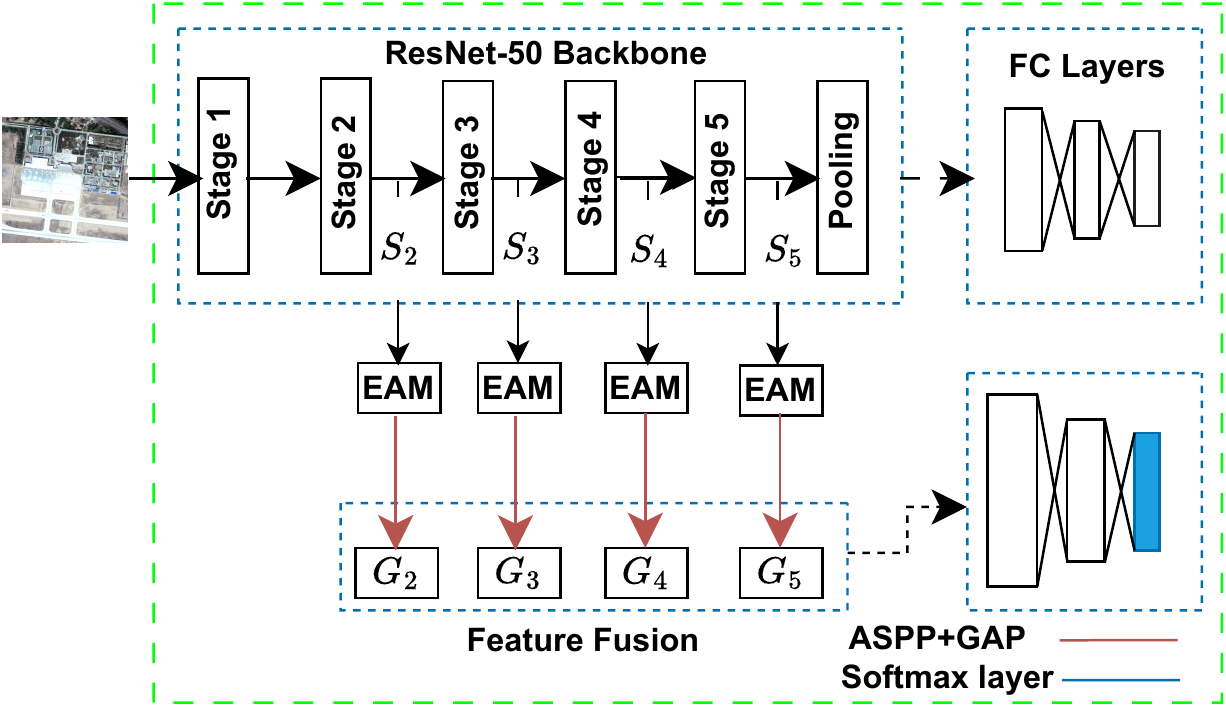}
  \caption{High-level pipeline of the proposed approach. Note that $S_2$, $S_3$, $S_4$, and $S_5$ denote the chosen layers from the second, third, fourth, and fifth stages, respectively. 
  }
  \label{fig:high_level}
\end{figure*}

\begin{figure*}
  \centering
  \includegraphics[height=85mm,width=0.77\textwidth, keepaspectratio]{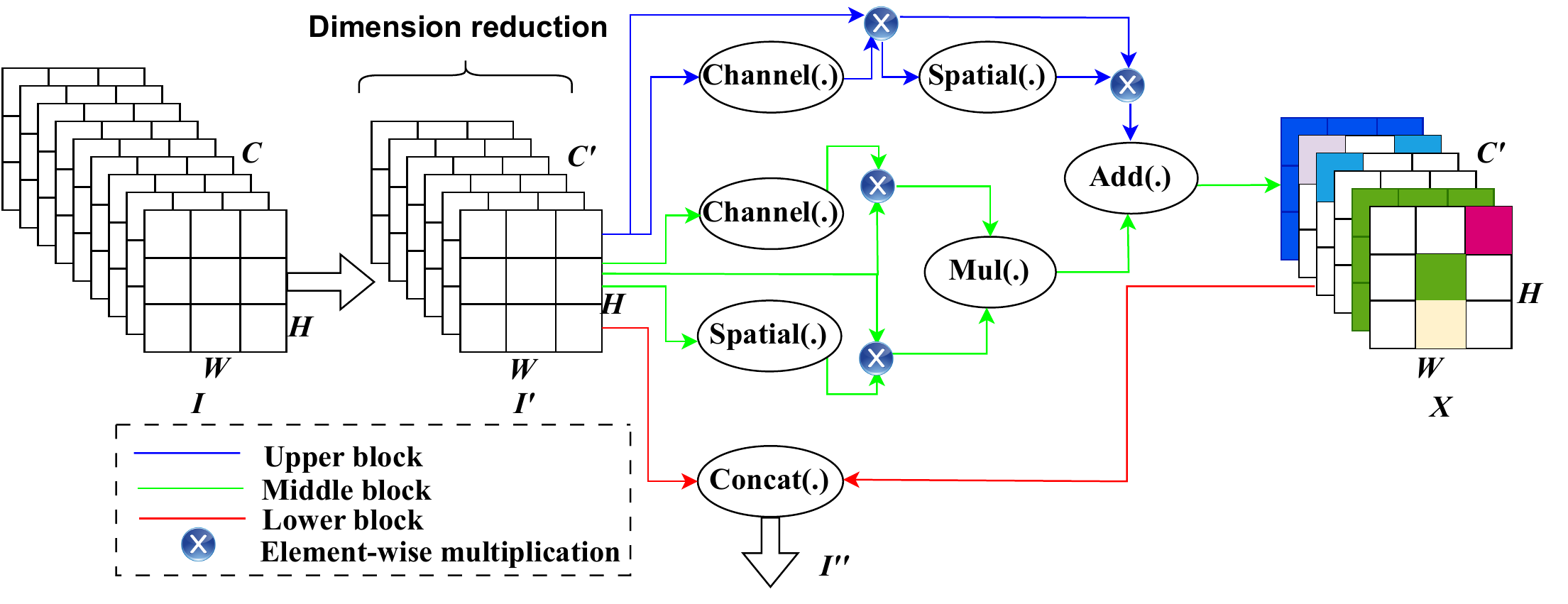}
  \caption{The pipeline of Enhanced VHR attention module (EAM). }
  \label{fig:EAM}
\end{figure*}

In the two-step approach, deep features are first extracted from the pre-trained DL models, also called transfer learning by feature extraction, and classified using a separate machine learning algorithm \cite{Weng2017Land-UseFeatures}.
Few seminal works have implemented the two-step approach compared to the end-to-end approach in the literature. For example, Weng et al. \cite{Weng2017Land-UseFeatures} extracted deep features using the AlexNet-like architecture pre-trained on the {\it ImageNet} and classified them using the extreme learning machine (ELM) classifier. Furthermore, Yu \& Liu \cite{Yu2018AClassification} extracted deep features from the VGG-16 and GoogleNet and classified them using the ELM classifier. Similarly, He et al. \cite{he2018remote} extracted the deep features from the pre-trained CNN models (AlexNet and VGG-16), stacked them, and calculated the covariance matrix to achieve the complementary information. After that, such features were used to classify using the support vector machine (SVM) classifier. 
Furthermore, Sun et al. \cite{sun2021multi} developed an approach to represent the image using both deep and hand-crafted features, which were then encoded using well-established methods such as bag-of-visual-words (BoVW) and scale-invariant scale transform (SIFT) for the classification using the SVM classifier. The involvement of innovative deep feature extraction and the appropriate encoding steps improved the overall classification performance. The two-step approach has provided an encouraging result; however, they require the expertise of multiple algorithms to get maximum benefits and still provide an unstable performance.

The existing DL-based methods produce an excellent overall classification performance; however, their results are still unstable or less generalisable. 
To deal with this shortcoming, we propose a novel DL-based approach, following the enhanced multi-level feature extraction procedure for higher discriminability considering the efficacy of multi-scale information as discussed by recent studies \cite{aryal2023multi,tavakkoli2019landslide}. 
Here, our approach first devises an enhanced VHR attention module (EAM), whose main component is the innovative spatial and channel attention-based mechanism inspired by the convolutional block attention module (CBAM) \cite{woo2018cbam}. 
Specifically, the proposed mechanism first extracts the new semantic information using parallel attention and then combines it with sequential attention. The original CBAM, which only uses serial attention, is unable to capture the parallel semantic information, whereas our improved version captures both serial and parallel rich semantic information.
In light of the efficacy of the multi-scale information to understand the input image at a finer level, we further leverage the atrous spatial pyramid pooling (ASPP) \cite{chen2018encoder,atrous}, which has a strong discriminative capability as suggested by Chen et al. \cite{chen2018encoder}, and finally aggregate using the global average pooling (GAP), preserving both maximum and minimum activation values for better representation for more accurate classification.

In summary, the main {\bf contributions} in this paper are as follows:

\begin{enumerate}
    \item[1)] We develop a novel EAM to attain the enriched salient information, which is appropriate to the aerial RS image datasets. 
    \item[2)] We propose an improved CBAM (ICBAM) as a backbone of the proposed EAM, which is an improved version of CBAM.
    \item [3)] We investigate the efficacy of various attention mechanisms within our proposed EAM for the aerial VHR RS scene classification.    
    \item [4)] We propose to use the ASPP over EAM-based features and aggregate them by the GAP operation (EAM+ASPP+GAP), producing the multi-scale salient information for finer-level information.
    \item [5)] \textcolor{red}{We show the explainability of the EAM using Gradient-weighted Class Activation Mapping (Grad-CAM) \cite{selvaraju2017grad} technique, which reveals comparatively larger discriminative regions during classification.}    
    \item[6)] We demonstrate the superiority of the proposed approach on two commonly-used VHR RS datasets through an extensive ablation and comparative study with the state-of-the-art (SOTA) methods.
\end{enumerate}
The article is organised as follows. Section "Proposed approach" elaborates the proposed approach in a step-wise manner. Similarly, Section "Experiment and analysis" presents the experimental settings, datasets, results and discussion. Lastly, Section "Conclusion" concludes the paper with future works/recommendations.


\section{Proposed approach}
\label{prop_app}
The proposed approach {(Figure \ref{fig:high_level})} for high-level workflow has three major steps:
Feature extraction, EAM, and Feature fusion. 

\subsection{Feature extraction}
\label{level_1}

We resort to the ResNet-50 model \cite{he2016deep}  for the feature extraction, as suggested by Wang et al. \cite{wang2022}. In the ResNet-50 model, same-sized feature maps generated from the layers are called stages. In total, there are five stages in this model. Among them, the last four stages are useful for VHR RS scene classification \cite{wang2022} as they jointly have the capability to achieve information from the basic to the higher level. So, we exploit only the last layers from each of the four stages, representing each level. We denote the features extracted from those layers as $S_2$, $S_3$, $S_4$, and $S_5$, which provides the number of channels as $256$, $512$, $1024$, and $2048$, respectively.

\subsection{EAM}
\label{level_2}

The proposed enhanced VHR attention module (EAM) has four main blocks: dimension reduction, upper, middle, and lower. The pipeline of the EAM is shown in Figure \ref{fig:EAM}.
\subsubsection{Dimension reduction block}
\label{dim_reduction}

The uniform-sized tensors are pivotal to mitigate the risk of information bias provided by the higher-sized tensors as these tensors outweigh the lower-sized tensors during the concatenation-based feature fusion.
To achieve the uniform-sized tensor, we perform the $1 \times 1$ convolution operation (refer to Eq. \eqref{eq:reduce_dimension}) and feed it into the three blocks (upper, middle, and lower blocks).
\begin{equation}
    I'=f^{1\times 1}{(I)}
    \label{eq:reduce_dimension},
\end{equation}
where $I' \in \mathbb{R}^{C'\times H \times W}$ ($C'\le C$), which is obtained from the input tensor $I \in \mathbb{R}^{C\times H \times W}$ using the 2D convolution ($\it f$) with super-scripted kernel size of $1 \times 1$.

\subsubsection{Upper block}
\label{upper}

The upper block achieves the salient information from channel and spatial regions using the CBAM \cite{woo2018cbam}, which comprises two sequentially-arranged blocks, namely the channel attention block (Eq. \eqref{eq:upper1}), and the spatial attention block (Eq. \eqref{eq:upper2}).
\begin{equation}
 F'= Channel(I') \otimes I',
    \label{eq:upper1}
\end{equation}

\begin{equation}
F''= Spatial(F') \otimes F',
    \label{eq:upper2}
\end{equation}

where $F'$, $F''$, and $\otimes$ denote the output of channel attention block ($Channel(.)$), spatial attention block ($Spatial(.)$), and element-wise multiplication, respectively. 
As an example for the input tensor $I'$, the channel attention block ($Channel(I')$) uses max pooling ($MaxP(I')$)  and average pooling ($AvgP(I')$) operations, and adds ($+$) them, followed by one hidden-layered multi-layer perception (MLP) for both operations and then the {\it Sigmoid} activation ($\sigma$) on them (Eq. \eqref{eq:channel}).

\begin{equation}
Channel(I')= \sigma(MLP(AvgP(I')+MLP(MaxP(I')))
\label{eq:channel}
\end{equation}

Whereas, for $F'$, the spatial attention block ($Spatial(F')$) employs the max pooling ($MaxP(F')$) and average pooling ($AvgP(F')$), with their concatenation ($;$), followed by the 2D convolution over $7 \times 7$ kernel size ($f^{7 \times 7}$) with the {\it Sigmoid} activation (Eq. \eqref{eq:spatial}).
\begin{equation}
    Spatial(F')= \sigma(f^{7 \times 7}([AvgP(F');MaxP(F'])
    \label{eq:spatial}
\end{equation}

The sequential arrangement of channel and spatial attention blocks lacks the salient information that is obtained from the parallel arrangement, which is the complementary information for the separability of categories. To this end, we add a middle block to complement the upper block, producing complementary semantic information, which is crucial for separability.
 
\subsubsection{Middle block}
\label{middle}

Here, we combine the channel attention output (Eq. \eqref{eq:middle1}) with spatial attention output (Eq. \eqref{eq:middle2}) using the element-wise multiplicative operation (Eq. \ref{eq:middle3}), followed by the elementwise addition operation with the upper block output (Eq. \eqref{eq:middle4}). This helps combine the outputs obtained from the sequential attention block with the ones from the parallel attention block. 

\begin{equation}
 \lambda= Channel(I') \otimes I',
    \label{eq:middle1}
\end{equation}

\begin{equation}
\beta= Spatial(I') \otimes I',
    \label{eq:middle2}
\end{equation}

\begin{equation}
 \delta= Mul(\lambda, \beta),
    \label{eq:middle3}
\end{equation}

\begin{equation}
X= Add(F'', \delta),
    \label{eq:middle4}
\end{equation}

where $\lambda$, $\beta$, $\delta$, and $X$ denote the output from channel attention, spatial attention, element-wise multiplication ($Mult(.)$), and element-wise addition operation ($Add(.)$), respectively. \textcolor{red}{The combination of the upper block and the middle block forms the ICBAM, which results in $X$ as shown in Eq. \eqref{eq:middle4}.}


\subsubsection{Lower block}
\label{lower}

The lower block is designed to combine the convolutional features with the attention-based features obtained from the middle block (refer to Eq. \eqref{eq:lower_block}). Note that the simple concatenation operator is used to combine here.

\begin{equation}
    I''=Concat(I',X)
    \label{eq:lower_block},
\end{equation}

where $I''\in \mathbb{R}^{2C'\times H \times W}$ denotes the result obtained from the concatenation ($Concat(.)$) of $I'$ with $X$. Attention-based features alone are insufficient to discriminate the images accurately during classification, so it is essential to include convolutional features additionally for performance improvement (refer to Section \ref{sec_convolution_ablation}). Further, the effectiveness of the convolutional features has also been underscored in the previous study \cite{sitaula2021attention}.
To this end, this block combines the attention features obtained from the middle blocks (ICBAM's output) with the convolutional features achieved from the dimension reduction block.

\begin{figure}[b]
  \centering
  \includegraphics[height=50mm,width=0.49\textwidth, keepaspectratio]{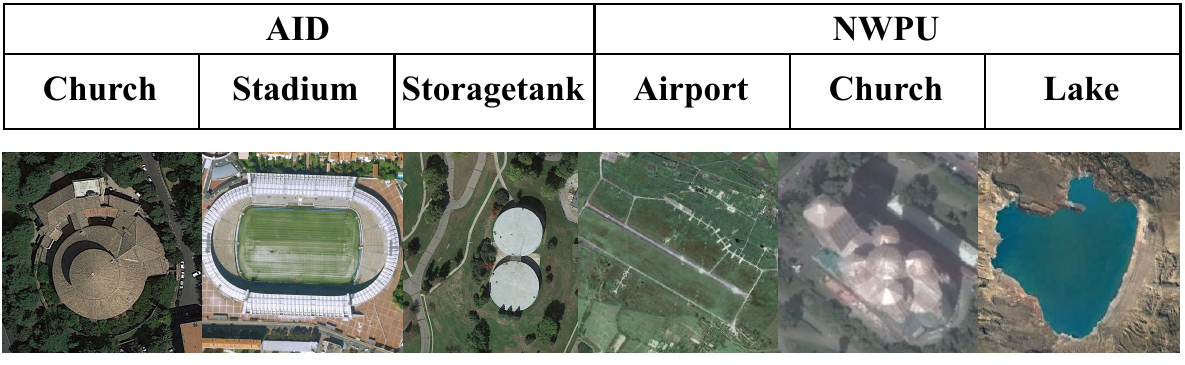}
  \caption{Sample images for three different categories belonging to AID and NWPU datasets}.
  \label{fig:sampleimage}
\end{figure}

\begin{figure}
\centering
\subfloat[AID]{\includegraphics[width=0.54\textwidth, height=47cm,keepaspectratio]{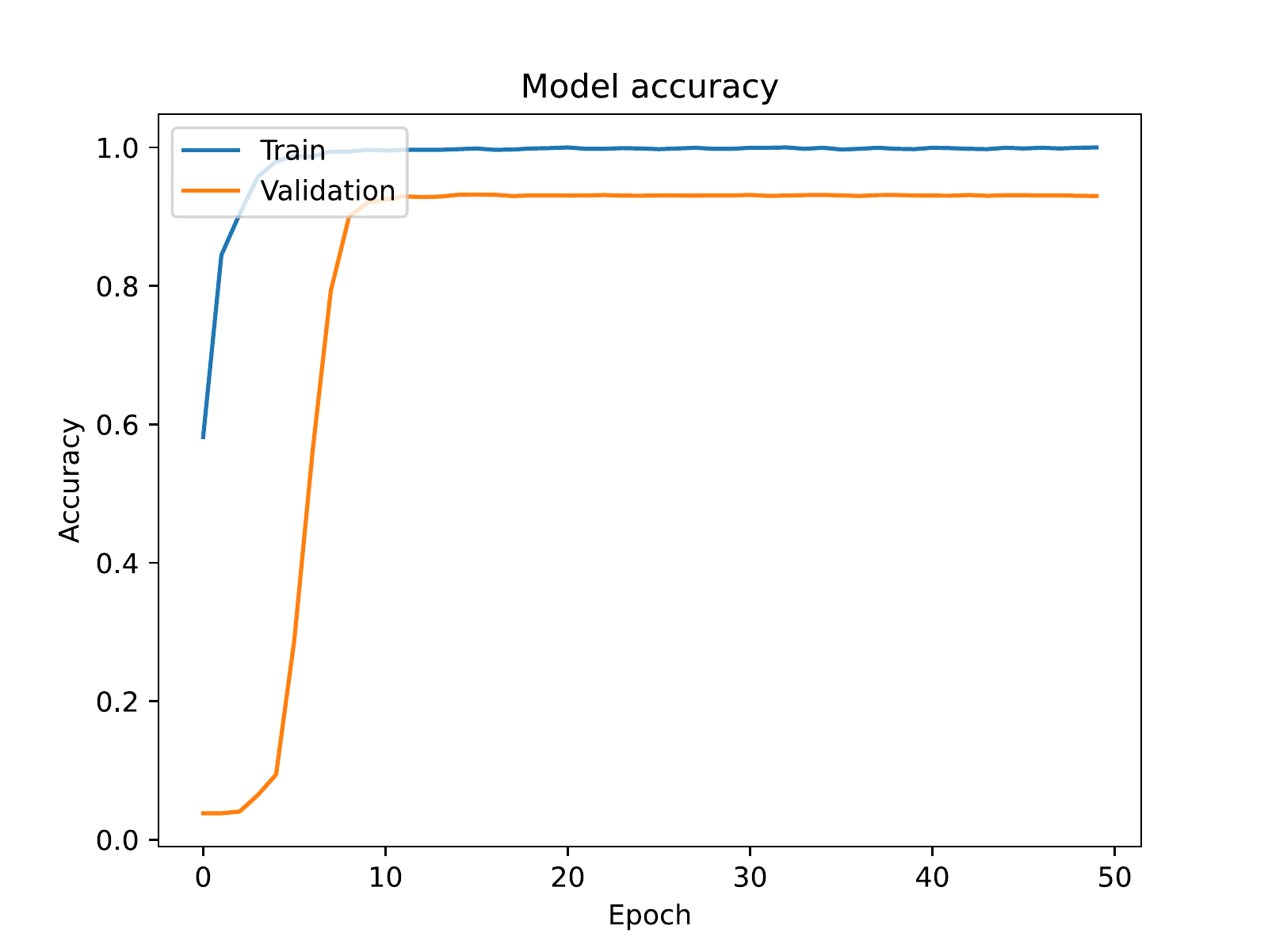}}\qquad
\subfloat[NWPU]{\includegraphics[width=0.54\textwidth, height=47cm,keepaspectratio]{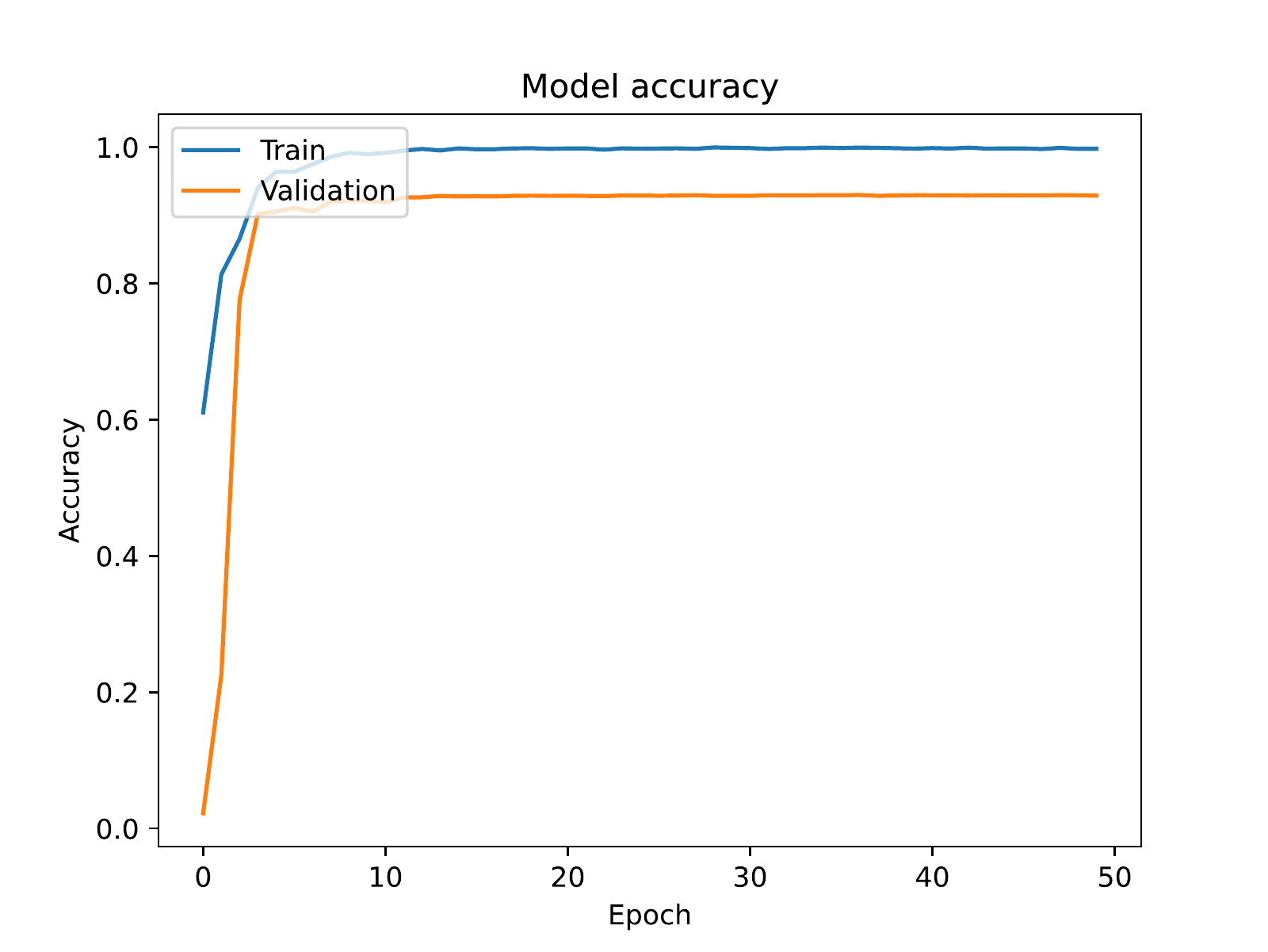}}\qquad
\caption{Training and validation curve produced while implementing our proposed approach on the AID (a) and NWPU (b) datasets.}
\label{fig:train_test_curve}
\end{figure}

\subsection{Feature fusion}
\label{level_3}

Let $g\colon S^{'}_{i}\rightarrow  S^{''}_{i}$, where $S^{'}_{i}$ and $S^{''}_{i}$ denote the input tensor extracted from the ResNet-50 and resultant output tensor achieved from the EAM (denoted by $g$ function here) for the $i^{th}$ layer (level), respectively. Note that $i\in \{2,\cdots,5\}$.
Such features are further enhanced using the ASPP operations to achieve the multi-scale information \cite{chen2018encoder} (Eq.\eqref{eq:aspp}).
Based on Chen et al. \cite{chen2018encoder}, we employ a kernel size of 1 with a dilation ratio of 1, and a kernel size of 3 with a dilation ratio of 6, 12, and 18 for the ASPP. 

\begin{equation}
    H_i = ASPP(S^{''}_i),
    \label{eq:aspp}
\end{equation}

where $H_i$ denotes the features obtained from the ASPP operation of the $i^{th}$ layer. 
Given that the ASPP produces tensors that need to be aggregated before fusion, we employ the global average pooling (GAP) operation, which not only provides the scale-invariant features   but also preserves both minimum and maximum activation information during aggregation (Eq. \eqref{eq:gap}).
\begin{equation}
    G_i=GAP(H_i),
    \label{eq:gap}
\end{equation}
where $G_i$ denotes the features obtained from GAP operation for the corresponding $i^{th}$ layer.
Finally, such features from all four layers are fused using a simple concatenation-based feature fusion approach (Eq. \eqref{eq:fusion}) and then classified into different classes using the {\it Softmax} layer.
\begin{equation}
    F=[G_2; G_3; G_4; G_5],
    \label{eq:fusion}
\end{equation}
where $F$ represents the enhanced multi-level features, which are used for the classification.

\begin{table*}[b]
\renewcommand{\arraystretch}{1.3}
\caption{Comparative study using overall classification accuracy (\%) $\,\pm\,$ standard deviation on the AID dataset. 
Please note that Tr=training, Te=testing, and the boldfaced values denote the highest performance, which is similar in the remaining tables.}
\centering
\begin{tabular}{p{4cm} p{3cm} p{3cm}}
    \hline\hline
    \textbf{Method} & 
    \textbf{Tr=50\% Te=50\%}& 
    \textbf{Tr=20\% Te=80\%}\\
    \hline
    ARCNet-VGG16 \cite{wang2018scene} & $93.10\,\pm\,0.550$ & $88.75\,\pm\,0.400$\\
    MSCP \cite{he2018remote} & $94.42\,\pm\,0.170$ & $91.52\,\pm\,0.210$\\
    CNN+ELM \cite{Yu2018AClassification} &$94.58 \pm\, 0.250$ &$92.32\, \pm\, 0.410$\\
    LCPB \cite{sun2021multi} & $91.33\,\pm\,0.360$ & $87.68\,\pm\,0.250$\\
    LCPP \cite{sun2021multi} & $93.12\,\pm\,0.280$ & $90.96\,\pm\,0.330$\\
    ORRCNN \cite{9178726} & $92.00$ & $86.42$\\
    ACR-MLFF \cite{wang2022} & $95.06\,\pm\,0.330$ & $92.73\,\pm\,0.120$\\
    MRHNet-50 \cite{Li2022} & $95.06$ &91.14\\
    {\bf EAM} & {$ \textBF{95.39\,\pm\,0.001}$} & {$\textBF{93.14\,\pm\,0.003}$}\\
    \hline\hline
\end{tabular}
\label{tab:AID}
\end{table*}

\begin{table*}[b] 
\renewcommand{\arraystretch}{1.3}
\caption{Comparative study using overall classification accuracy (\%) $\,\pm\,$ standard deviation on the NWPU dataset. Here, '-' denotes the unavailable performance. }
\centering
\begin{tabular}{p{3cm} p{3cm} p{3cm}}
    \hline\hline
    \textbf{Method}& 
    \textbf{Tr=10\% Te=90\%}& 
    \textbf{Tr=20\% Te=80\%}\\
    \hline
    MSCP \cite{he2018remote} & $88.07\,\pm\,0.180$ & $90.81\,\pm\,0.130$\\
    SCCov \cite{he2019skip} & $89.30\,\pm\,0.350$ & $92.10\,\pm\,0.250$\\
    SKAL \cite{9298485} & {$\textBF{90.41}$}$\,\pm\,0.120$ & $92.95\,\pm\,0.090$\\
    MRHNet-101 \cite{9431581} & - & 91.64\\
    ACR-MLFF  \cite{wang2022} & $90.01\,\pm\,0.330$ & $92.45\,\pm\,0.200$\\
    {\bf EAM} & $90.38\,\pm\,\textBF{0.001}$ & {$\textBF{93.04\,\pm\,0.001}$}\\
    \hline\hline
\end{tabular}
\label{tab:NWPU}
\end{table*}

\section{Experiment and analysis}
\label{experimental_analysis}

In this section, we discuss dataset specifications, experimental implementations and result analysis, followed by the ablation study.

\subsection{Datasets}

We perform our experiment on two commonly-used VHR RS datasets: Aerial Image Data Set (AID) \cite{xia2017aid}, and NWPU-RESISC45 (NWPU) \cite{cheng2017remote}. The AID contains a collection of 10,000 images divided into 30 classes. The number of images in each category varies from 220 to 420. Each image is $600 \times 600$ pixels in size, and the spatial resolution ranges from 0.5 to 8 m. Similarly, the NWPU is a larger RS dataset containing 31,500 images distributed into 45 classes. Each image has the size of $256 \times 256$ pixels and varying spatial resolution from 0.2 to 30 m. 
The sample images (third row) of AID and NWPU datasets are presented in Figure \ref{fig:sampleimage}.

For the evaluation, we design train:test splits using the standard method as in existing works \cite{wang2022,Li2022} for both datasets. For the AID dataset, we design two sets of five random train:test splits with a ratio of 50:50 and 20:80, respectively. Similarly, for the NWPU dataset, we prepare two sets of five random train:test splits with a ratio of 10:90 and 20:80, respectively. We report the averaged overall classification performance over five splits with the standard deviation for each set on both datasets. The sample training and validation curves generated by our proposed method, which shows a good fit, on both datasets are shown in Figure \ref{fig:train_test_curve}.


\begin{table*}[tb]
\renewcommand{\arraystretch}{1.3}
\caption{{Ablation study of three components using overall classification accuracy (\%) $\,\pm\,$ standard deviation on both datasets.}}
\centering
\begin{tabular}{p{3.3cm} p{2.6cm} p{2.6cm}p{2.6cm} p{2.6cm}}
    \hline\hline
      \textbf{Strategy}&
      \multicolumn{2}{c|}{\textbf{NWPU}}&
      \multicolumn{2}{c}{\textbf{AID}} \\
      \cline{2-5}
    & \textbf{Tr=10\% Te=90\%} & 
    \textbf{Tr=20\% Te=80\%}&
    \textbf{Tr=50\% Te=50\%} & 
    \textbf{Tr=20\% Te=80\%}\\
    \hline
    GAP& {$90.04\,\pm\,0.003$} &{$92.88\,\pm\,0.002$}&{$95.04\,\pm\,0.004$}&{$92.63\,\pm\,\textBF{0.001}$}\\ 
    ASPP + GAP& {$89.99\,\pm\,0.003$} &{$92.88\,\pm\,\textBF{0.001}$}&{$95.30\,\pm\,0.002$}&{$92.76\,\pm\,0.004$}\\ 
    EAM + GAP& {$89.52\,\pm\,0.004$} &{$92.64\,\pm\,0.002$}&{$94.70\,\pm\,0.003$}&{$92.02\,\pm\,0.004$}\\
    \bf{EAM + ASPP + GAP}& {$\textBF{90.38\,\pm\,0.001}$} &{$\textBF{93.04\,\pm\,0.001}$}&{$\textBF{95.39\,\pm\,0.001}$}&{${\textBF{93.14}\,\pm\,0.003}$}\\
    \hline
    \hline
\end{tabular}
\label{tab: ablative}
\end{table*}

\begin{table*}[tb]
\renewcommand{\arraystretch}{1.3}
\caption{{Comparative study of various attention mechanisms with EAM using overall classification accuracy (\%) $\,\pm\,$ standard deviation on both datasets.}}
\centering
\begin{tabular}{p{3.5cm} p{2.6cm} p{2.6cm}p{2.6cm} p{2.6cm}}
    \hline\hline
      \textbf{Attention mechanism}&
      \multicolumn{2}{c|}{\textbf{NWPU}}&
      \multicolumn{2}{c}{\textbf{AID}} \\
      \cline{2-5}
    & \textbf{Tr=10\% Te=90\%} & 
    \textbf{Tr=20\% Te=80\%}&
    \textbf{Tr=50\% Te=50\%} & 
    \textbf{Tr=20\% Te=80\%}\\
    \hline
    CBAM& {$90.36\,\pm\,0.263$} &{$92.90\,\pm\,0.220$}&{$94.79\,\pm\,0.851$}&{$92.81\,\pm\,{0.003}$}\\ 
    SENet& {$\textBF{90.50}\,\pm\,0.002$} &{$\textBF{93.20\,\pm\,{0.001}}$}&{$95.37\,\pm\,0.320$}&{$93.03\,\pm\,0.001$}\\ 
    SC-SENet& {$90.37\,\pm\,0.002$} &{$93.10\,\pm\,{0.002}$}&{$95.29\,\pm\,0.001$}&{$92.88\,\pm\,0.001$}\\ 
    DualNet& {$90.28\,\pm\,0.002$} &{$92.95\,\pm\,{0.001}$}&{$95.39\,\pm\,0.001$}&{$92.96\,\pm\,0.001$}\\ 
    {\bf ICBAM}& {${90.38\,\pm\,\textBF{0.001}}$} &{$93.04\,\pm\,\textBF{0.001}$}&{$\textBF{95.39\,\pm\,0.001}$}&{$\textBF{93.14}\,\pm\,0.003$}\\ 
    \hline
    \hline
\end{tabular}
\label{tab: ablative_attention}
\end{table*}

\begin{figure*}
  \centering
  \includegraphics[height=70mm,width=0.95\textwidth, keepaspectratio]{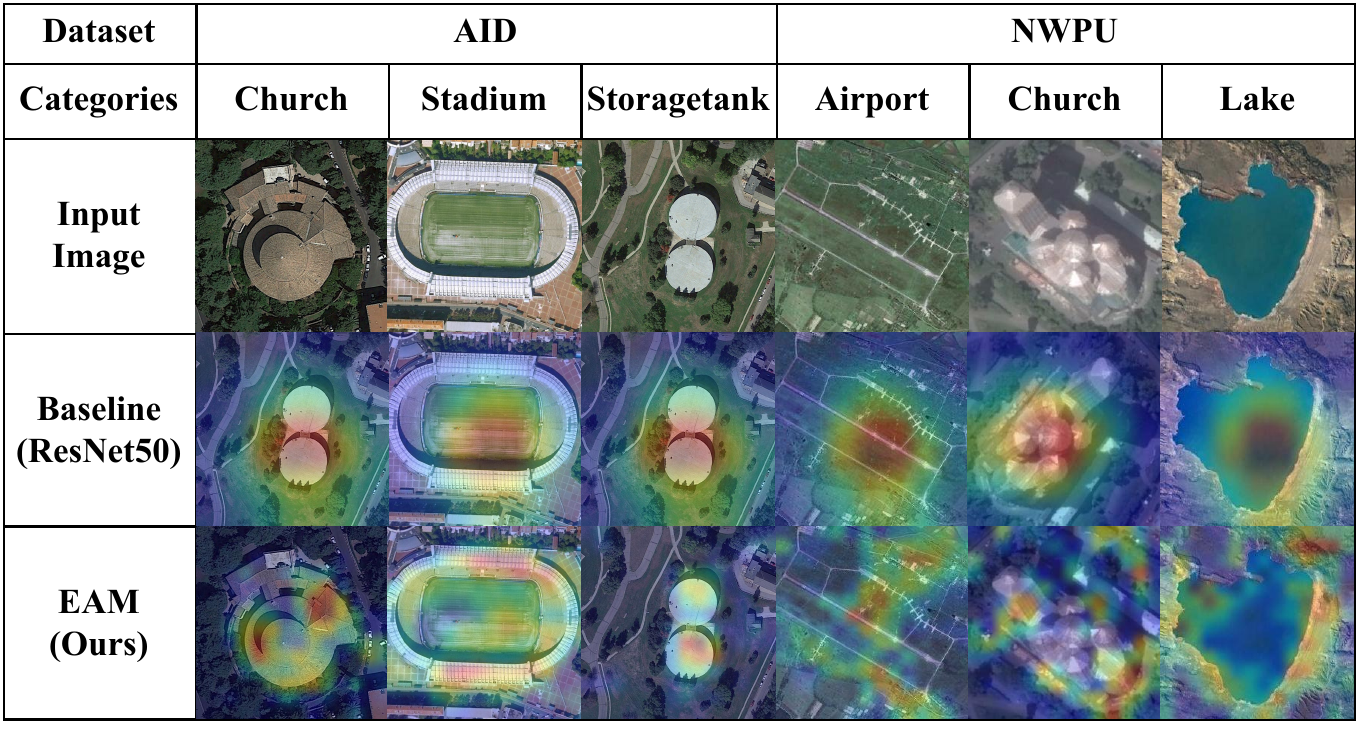}
  \caption{Visualisation of discriminative regions produced by the ResNet-50 and our proposed method (EAM) with the help of Grad-CAM visualisation \cite{selvaraju2017grad}. }
  \label{fig:featuremap}
\end{figure*}


\begin{figure*}[!htbp]
  \centering
  \includegraphics[height=300mm,width=0.90\textwidth, keepaspectratio]{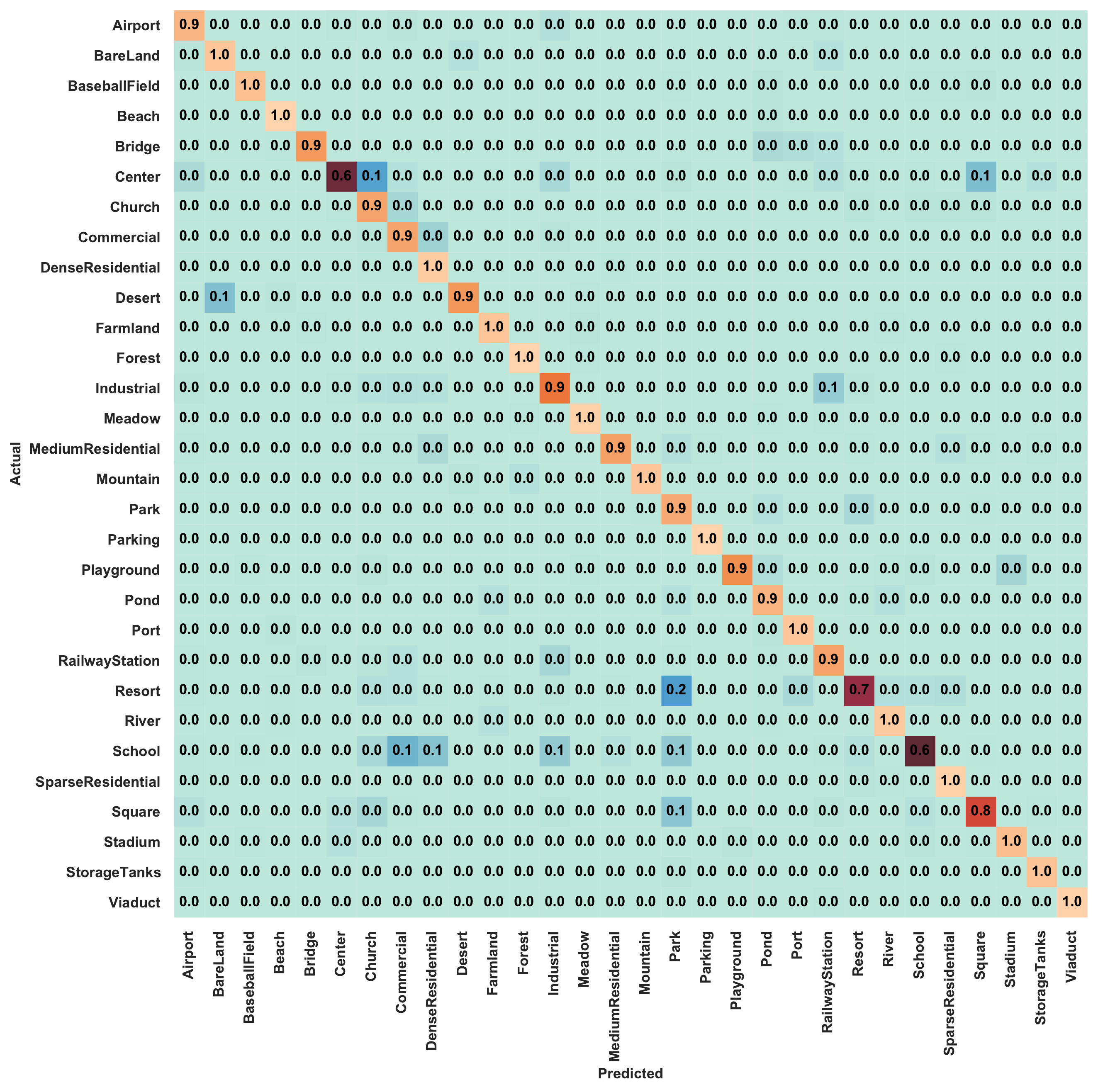}
  \caption{Category-wise performance analysis on the AID dataset. }
  \label{fig:confusion_AID}
\end{figure*}

\begin{figure*}[!htbp]
  \centering
  \includegraphics[height=460mm,width=0.99\textwidth, keepaspectratio]{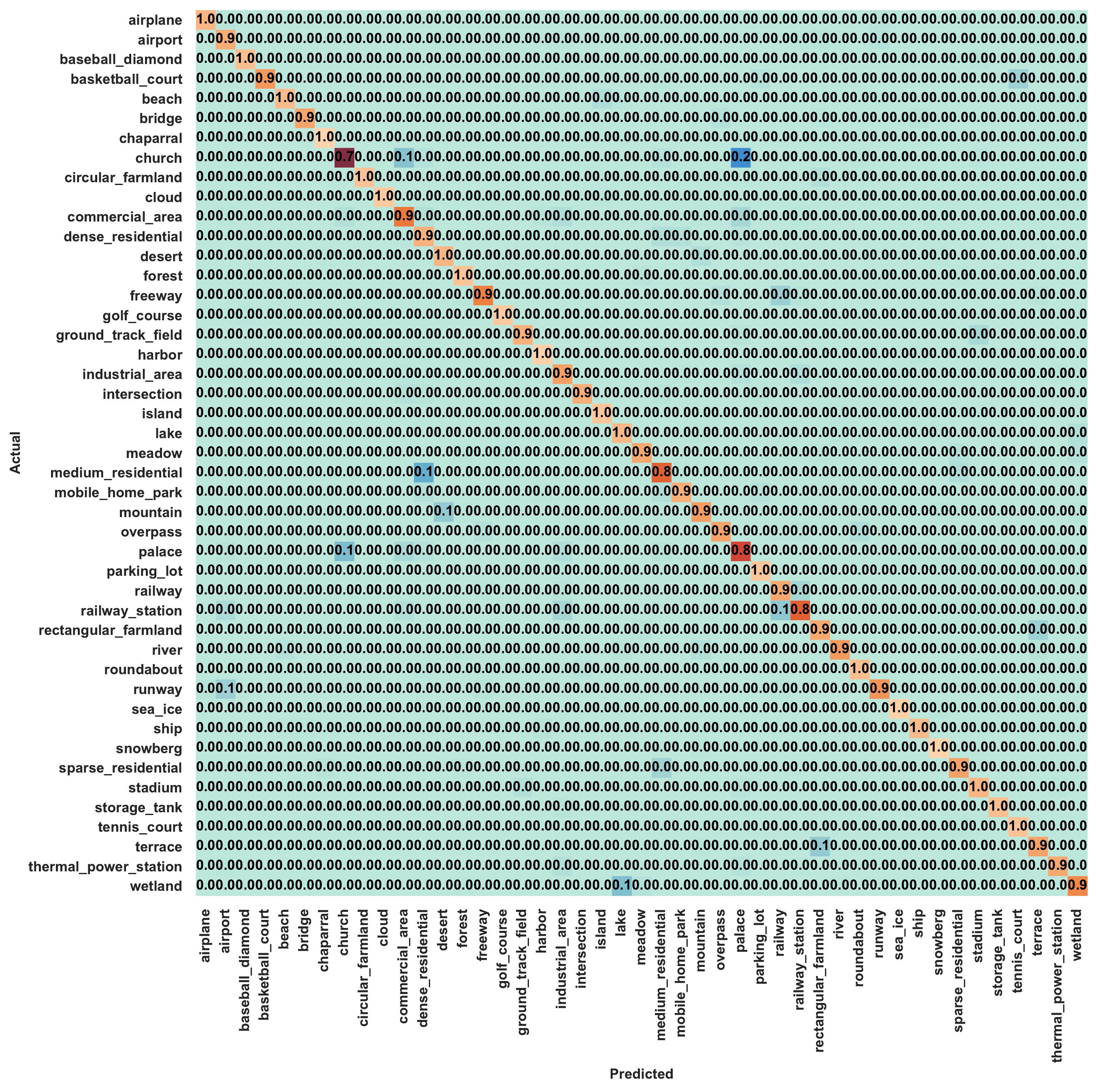}
  \caption{Category-wise performance analysis on the NWPU dataset.} 
  \label{fig:confusion_NWPU}
\end{figure*}

\subsection{Implementation}
All experiments are performed using NVIDIA A100 GPU (Graphical Processing Unit) with the Keras under Tensorflow package \cite{abadi2016tensorflow} in Python. We employ the {\it Adam} optimizer, where we set the weight decay penalty and batch\_size to {$1 \times 10^{-4}$} and {16}, respectively. The learning rate and the number of epochs are set to $3 \times 10^{-3}$ and 50, respectively. We apply a horizontal flip with random cropping and shuffling operations for data augmentation. 

\begin{table*}[]
\centering
\renewcommand{\arraystretch}{1.3}
\caption{Comparative study of with (w/) and without (w/o) convolutional features in EAM 
using overall classification accuracy (\%) $\,\pm\,$ standard deviation on both datasets.}
\begin{tabular}{p{2cm}p{3cm}p{2cm}p{2cm}}
\hline
\hline
\textbf{Dataset} & \textbf{Setting} & \textbf{w/ }& \textbf{w/o} \\
\hline
\multirow{2}{*}{AID}  & Tr=20\% Te=80\%     &  {$\textBF{93.14\,\pm\,0.003}$}  &{$91.46\,\pm\,0.290$}     \\
& Tr=50\% Te=50\%     & {$\textBF{95.39\,\pm\,0.001}$}   &{$94.71\,\pm\,0.004$}     \\
                   
\hline
\multirow{2}{*}{NWPU} & Tr=20\% Te=80\%     &{$\textBF{93.04\,\pm\,0.001}$}    &   {$92.40\,\pm\,0.001$}  \\
    & Tr=10\% Te=90\%     &{$\textBF{90.38\,\pm\,0.001}$}    &{$89.20\,\pm\,0.480$}    \\
\hline
\hline
\end{tabular}
\label{tab:with_without_convolutional}
\end{table*}

\subsection{Results and discussion}
\label{results}
\subsubsection{Comparison with the SOTA methods}

We {compare} the overall classification accuracy of our proposed approach with the SOTA methods on both {the} datasets.
The results are presented in Tables \ref{tab:AID} and \ref{tab:NWPU} for the AID and NWPU datasets, respectively.
From Table \ref{tab:AID}, we observe that our approach outperforms the recent methods for 50:50 and 20:80 settings with the stable classification accuracy of 95.39\% (0.33\% higher than the best-performing method (ACR-MLFF \cite{wang2022})), and 93.14\% (0.41\% higher than the best-performing method, ACR-MLFF \cite{wang2022}), respectively. We also observe that it shows stable performance with the lowest standard deviation of 0.001 for the 50:50 setting and 0.003 for the 20:80 setting. 

From Table \ref{tab:NWPU}, we notice that our approach yields prominent classification accuracy of 90.38\% for the 10:90 setting (0.03\% lower than the best-performing method, SKAL \cite{9298485}) and 93.04\% for the 20:80 setting (0.09\% higher than the best-performing method, SKAL \cite{9298485}). Although the proposed approach imparts a slightly lower accuracy than the best-performing method for the 10:90 setting on the NWPU, it shows consistency/robustness across performance measures with the lowest standard deviation (0.001 for both the 10:90 and 20:80 settings).

The comparative study with SOTA methods on both datasets reveals that our proposed method yields improved performance with the least standard deviation. Such encouraging results are attributed to the highly discriminating multi-level attention information extracted by our method during the classification. 

\subsubsection{Ablation study of three components}

To understand the role of EAM, ASPP, and GAP on all four layers, we analyse the performance of their four sequential combination strategies: GAP, ASPP+GAP, EAM+GAP, and EAM+ASPP+GAP. Note that GAP is used in every combination to achieve the final aggregated results. The results are presented in Table \ref{tab: ablative}. 
From Table \ref{tab: ablative}, we find that the sequential combination of three components (EAM+ASPP+GAP) provides more stable and improved performance compared to the remaining strategies. This result is attributed to the joint involvement of all three sequential components.

\subsubsection{EAM analysis with various attention mechanisms}
To investigate the efficacy of commonly-used attention mechanisms (CBAM \cite{woo2018cbam}, SENet(Squeeze Excitation Networks) \cite{hu2018squeeze}, spatial channel (SC)-SENet \cite{roy2018concurrent}, DualNet \cite{fu2019dual,dual}) including ours (ICBAM) with the proposed EAM, we plug each of them within the EAM block and perform the comparison using overall performance and standard deviation on both datasets. For this, we replace the upper and middle blocks of the EAM with the aforementioned attention mechanisms with each attention mechanism. We report the comparative results in Table \ref{tab: ablative_attention}. 

From Table \ref{tab: ablative_attention}, we find that the EAM with ICBAM has a highly discriminative feature extraction ability, evidenced mainly by the improved performance with stability than other counterparts on both datasets. 
However, there is an exception in the 20: 80 setting of the NWPU dataset, where SENet outperforms ours in terms of overall performance and stability. 
Further, it is also worth noting that our method delivers a more stable performance than counterparts in the 10: 90 setting on the NWPU dataset although it has a slightly lower performance compared to the SENet mechanism. Overall, the addition of ICBAM with EAM outperforms other remaining counterparts in terms of competent performance with stability. 

\subsubsection{Qualitative analysis}

We perform a qualitative analysis using the feature map visualisation technique in this study. For this, the visual feature maps are achieved using the Grad-CAM \cite{selvaraju2017grad} and presented in Fig. \ref{fig:featuremap}.
From Figure \ref{fig:featuremap}, we find that our proposed method is capable of capturing more discriminative semantic regions than the baseline model (ResNet-50) during classification on both datasets. 
For example, for the image from the Church category on the AID dataset, we notice that the baseline provides only one discriminative region (reddish yellow) ignoring the peripheral buildings, whereas our proposal identifies additional reddish yellow spots representing the peripheral buildings. We observe a similar case for the image from other categories such as Church and Lake on the NWPU dataset.
Here, the identification of more discriminative regions is attributed to the multi-scale attention information captured by our proposal.

\subsubsection{Convolutional features analysis}
\label{sec_convolution_ablation}

To understand the contribution of convolutional features in our proposed EAM, we design and analyse two scenarios: i) with convolutional features (w/) and ii) without convolutional features (w/o). The comparative results on both datasets are presented in Table \ref{tab:with_without_convolutional}. 
The result shows that the w/ scenario performs better than the w/o scenario in each setting on both datasets. To this end, we believe that the convolutional features are as important as the attention features in our proposed EAM to improve the overall performance.

\subsubsection{Category-wise performance analysis}
\label{class_wise}

We perform the category-wise performance analysis of the proposed method on both datasets using the confusion matrix.
Figures \ref{fig:confusion_AID} and \ref{fig:confusion_NWPU} present the category-wise matrix representing the AID and NWPU, respectively.
From Figures \ref{fig:confusion_AID} and \ref{fig:confusion_NWPU}, we find that the accuracy of most categories is above or equal to 90\% on both datasets except for a few categories (such as School and Resort from AID dataset).
This underlines that the features delivered by our proposed approach are well-discriminating among categories, thereby imparting an excellent category-wise performance.

\section{Conclusion}
\label{conclusion}

In this paper, we propose a novel DL-based approach, called EAM, to represent and classify VHR RS images more accurately and consistently.
Our approach, which delivers a richer discriminative multi-scale salient information at a finer level, imparts excellent performance {with the highest accuracy of 95.39\% on the AID and 93.04\% on the NWPU datasets, where both of them have the least standard deviation (0.001).} 
For future work, we recommend working on the proposed EAM with various pre-trained DL models (such as VGG-16, Xception, and EfficientNet) on such datasets. Additionally, the potentiality of the proposed EAM and CBAM could be explored to boost the classification {using different publicly available and benchmark datasets.}

\section{Acknowledgement}
This research was supported by The University of Melbourne’s Research Computing Services and the Petascale Campus Initiative.
Also, the pre-print version of this manuscript is available on arXiv \cite{sitaula2023enhanced}.

\section{Authorship contribution statement}

C Sitaula: Data Curation, Conceptualization, Methodology, Software, writing, original draft preparation, Writing-review and editing. S KC: Conceptualization, Methodology, Writing-review and editing, 
J Aryal: Supervision, Validation, Resources, and Project Administration.

\section{Declaration of competing interest}
The authors have no competing interests to declare that are relevant to the content of this article.

\section{Data availability}
All datasets used in this paper are publicly available.

\section{Code availability}
The source code will be made available upon request.
The code will be available at xxxx<<github link>>>


\bibliographystyle{elsarticle-num}
\bibliography{references_.bib}
\end{document}